\documentclass[10pt,twocolumn,letterpaper]{article}

\pdfoutput=1

\usepackage{cvpr}              
\usepackage[dvipsnames]{xcolor}

\definecolor{cvprblue}{rgb}{0.21,0.49,0.74}
\usepackage[pagebackref,breaklinks,colorlinks,citecolor=cvprblue]{hyperref}

\newcommand{\modelname}{\textcolor{black}{ProbTalk}\xspace}

\definecolor{DeltaColor}{rgb}{0.039,0.73,0.71}
\definecolor{SigmaColor}{rgb}{0.98,0.45,0.0}
\definecolor{AlphaColor}{rgb}{0,0,0.8}
\definecolor{BetaColor}{rgb}{0.8,0,0.8}
\definecolor{GammaColor}{rgb}{0.514,0.34,0.224}
\definecolor{EpsilonColor}{rgb}{0.353,0.725,0.906}
\definecolor{PurpleColor}{rgb}{0.5,0,0.7}
\definecolor{OrangeColor}{rgb}{0.914,0.541,0.141}
\definecolor{GreenColor}{rgb}{0.137,0.573,0.565}
\definecolor{RedColor}{rgb}{0.949,0.275, 0.224}
\definecolor{LightCyan}{rgb}{0.88,1,1}
\definecolor{Gray}{gray}{0.3}
\definecolor{Strawberry}{rgb}{1,0.26,0.64}

\definecolor{BetaColor}{rgb}{0.8,0,0.8}

\definecolor{LightCyan}{rgb}{0.88,1,1}
\definecolor{lightgray}{rgb}{0.9,0.9,0.9}

\newcommand{\qheading}[1]{\noindent\textbf{#1}}
\newcommand{\tabincell}[2]{\begin{tabular}{@{}#1@{}}#2\end{tabular}} 

\definecolor{GreenColor}{rgb}{0.137,0.573,0.565}

\newcommand{\projectURL}{\url{https://feifeifeiliu.github.io/probtalk/}}

\renewcommand{\paragraph}[1]{\medskip\noindent\textbf{#1}\ \ }

\makeatletter
\newcommand*\bigcdot{\mathpalette\bigcdot@{.5}}
\newcommand*\bigcdot@[2]{\mathbin{\vcenter{\hbox{\scalebox{#2}{$\m@th#1\bullet$}}}}}
\makeatother

\usepackage{color}
\usepackage{amsthm}
\usepackage{wrapfig}
\usepackage{array}
\usepackage{newlfont} %
\usepackage{textcomp}
\usepackage{multirow}
\usepackage{hhline}
\usepackage{tabularx}
\usepackage{algorithmic}
\usepackage{microtype}
\usepackage{mathtools}
\usepackage{booktabs} %
\usepackage[ruled]{algorithm2e} %
\usepackage{comment}
\usepackage{times}
\usepackage{epsfig}
\usepackage{graphicx}
\usepackage{amsmath}
\usepackage{cuted}
\usepackage{capt-of}
\usepackage{enumitem}
\usepackage{balance}
\usepackage[normalem]{ulem} 
\usepackage[tone, extra, safe]{tipa}
\usepackage{tipx}
\usepackage{amssymb}
\usepackage{enumitem}
\usepackage{svg}
\usepackage[dvipsnames]{xcolor}
\usepackage{arydshln}
\usepackage[toc,page]{appendix}
\usepackage[accsupp]{axessibility}

\def\paperID{6012} 
\def\confName{CVPR}
\def\confYear{2024}

\title{Towards Variable and Coordinated Holistic Co-Speech Motion Generation} 

\begin{document}
\author{Yifei Liu$^{1,2*}$\quad Qiong Cao$^{2*}$\quad Yandong Wen$^{3}$\quad 
 Huaiguang Jiang$^{1}$\quad Changxing Ding$^{1\dagger}$ \\
$^{1}$South China University of Technology ~~~ $^2$JD Explore Academy \\
$^{3}$Max Planck Institute for Intelligent Systems, T\"ubingen, Germany \\
{\tt\small ft\_lyf@mail.scut.edu.cn, mathqiong2012@gmail.com} \\
{\tt\small yandong.wen@tuebingen.mpg.de, \{hihuagong2021, chxding\}@scut.edu.cn} \\
}



\captionsetup[figure]{hypcap=false}

\newcommand{\teasercaption}{\textbf{Holistic co-speech motion generation examples.} Given a speech signal as input, our approach generates variable and coordinated holistic body motions. From top to bottom: the speech transcript, the corresponding audio, and three generated samples.
In particular, to emphasize important keywords, our method ensures that facial expressions, head movements, and body motions work in unison.

}

\twocolumn[{
    \renewcommand\twocolumn[1][]{#1}
    \vspace{-0.5cm}
    \maketitle
    \centering
    
    \begin{minipage}{1.00\textwidth}
        \centering
        \vspace{-0.8cm}
        \includegraphics[width=\textwidth]{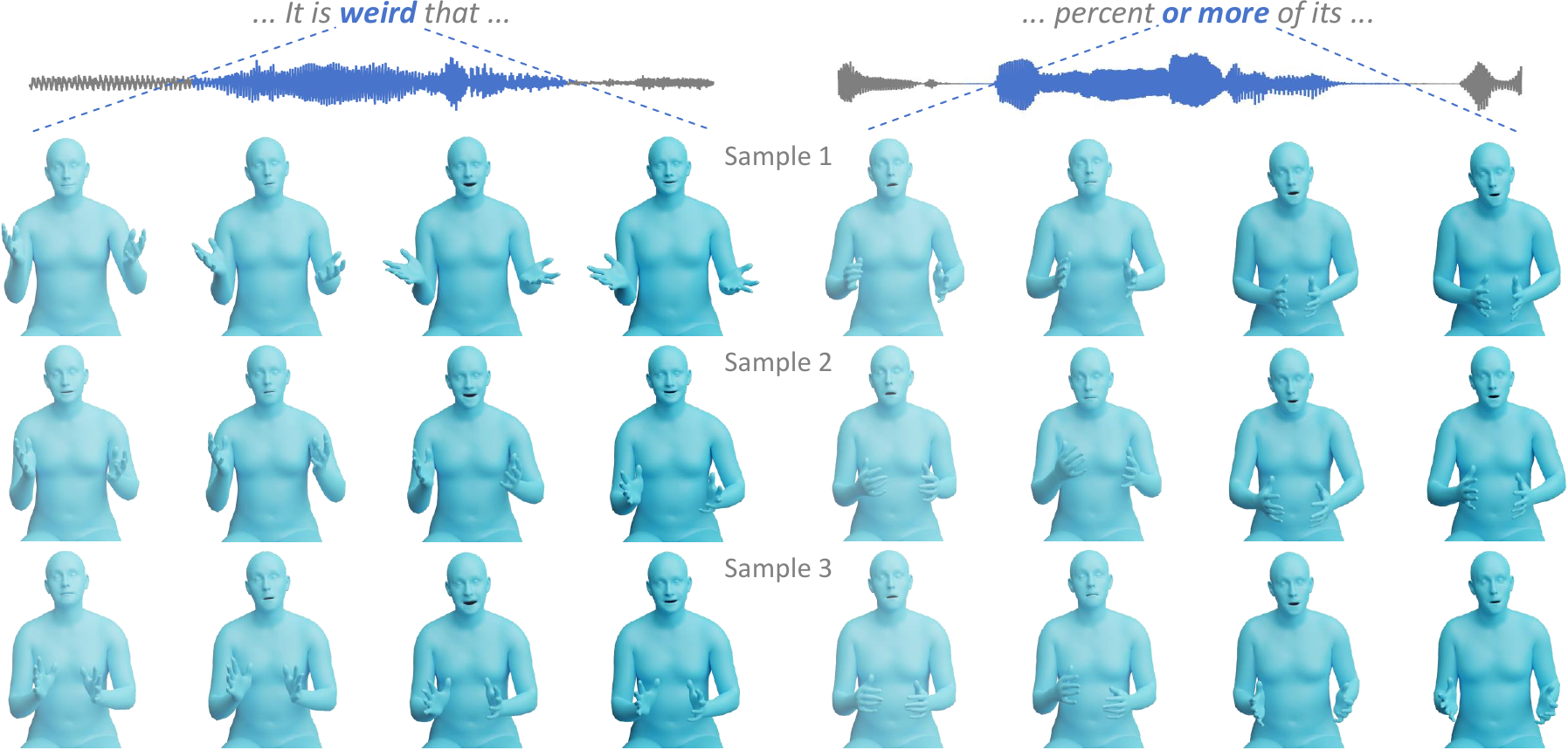}
    \end{minipage}
    \captionof{figure}{\teasercaption}
    \vspace{0.5cm}
    \label{fig:teaser}
}]

\def\thefootnote{*}\footnotetext{Equal Contribution.}
\def\thefootnote{$\dagger$}\footnotetext{Corresponding Author.}
\begin{abstract}

This paper addresses the problem of generating lifelike holistic co-speech motions for 3D avatars, focusing on two key aspects: variability and coordination. Variability allows the avatar to exhibit a wide range of motions even with similar speech content, while coordination ensures a harmonious alignment among facial expressions, hand gestures, and body poses. We aim to achieve both with \modelname, a unified probabilistic framework designed to jointly model facial, hand, and body movements in speech. \modelname builds on the variational autoencoder (VAE) architecture and incorporates three core designs. First, we introduce product quantization (PQ) to the VAE, which enriches the representation of complex holistic motion. Second, we devise a novel non-autoregressive model that embeds 2D positional encoding into the product-quantized representation, thereby preserving essential structure information of the PQ codes.
Last, we employ a secondary stage to refine the preliminary prediction, further sharpening the high-frequency details. Coupling these three designs enables \modelname to generate natural and diverse holistic co-speech motions, outperforming several state-of-the-art methods in qualitative and quantitative evaluations, particularly in terms of realism.
Our code and model will be released for research purposes at \projectURL.

\end{abstract}
    
\section{Introduction}

Studies in psychology and linguistics suggest that communication is not just about what we hear; it is a comprehensive sensory experience integrating non-verbal signals like body poses, hand gestures, and facial expressions, all crucial to effective communication \cite{goldin1999role, kendon2004gesture}. Consequently, the automatic generation of realistic full-body movements synchronized with speech is vital for offering more immersive and interactive user experiences.

This problem has seen considerable research interest. An early approach by Habibie et al. \cite{habibie2021learning} mapped speech signals to holistic motions using a deterministic regression model. While effective in some respects, this method risked producing repetitive and less human-like avatars due to identical motions for the same speech content. To improve on this, TalkSHOW \cite{yi2023generating} introduced a hybrid approach, using deterministic modeling for facial expressions and probabilistic modeling (specifically, VQ-VAE \cite{van2017neural}) for hand and body movements. Despite achieving a greater variety in body gestures, TalkSHOW still suffers from the low diversity in facial movements. Moreover, the separate modeling strategy used in TalkSHOW can cause disjointed coordination among different body parts.

Recent studies have shown the efficacy of probabilistic models like VQ-VAE in learning diverse facial movements \cite{ng2022learning, xing2023codetalker} or body gestures \cite{ao2022rhythmic, liu2022learning, liang2022seeg}, suggesting a promising direction for holistic motion modeling. Inspired by their successes, we aim to develop a variational framework that jointly models facial expressions, hand gestures, and body poses from speech. This framework is designed to capture both the variability and holistic coordination in co-speech motion. Specifically, the probabilistic nature introduces essential variability to the resulting motions, allowing avatars to exhibit a wide range of movements for similar speech, while the joint modeling improves the coordination, encouraging a harmonious alignment across various body parts. Yet, developing such a framework is not straightforward. Directly using models like VQ-VAE results in inexpressive speaking avatars, as holistic motion has a higher complexity than that of individual body parts. Moreover, the temporal granularity within VQ-VAE is typically too coarse \cite{ao2022rhythmic, yi2023generating}, constraining its ability to generate high-frequency motions with necessary details. 

To address these challenges, we present \modelname, a novel framework based on the variational autoencoder (VAE) architecture, comprising three core designs. First, we apply product quantization (PQ) to VAE \cite{gray1984vector, wu2019learning}. PQ partitions the latent space of holistic motion into multiple subspaces for individual quantization. The compositional nature of PQ-VAE affords a richer representation so that the complex holistic motion can be represented with lower quantization errors. Second, we devise a novel non-autoregressive model that integrates MaskGIT \cite{chang2022maskgit} and 2D positional encoding into PQ-VAE. MaskGIT \cite{chang2022maskgit} is a training and inference paradigm that simultaneously predicts all latent codes, significantly reducing the steps required for inference. The 2D positional encoding, accounting for the additional dimension introduced by PQ, effectively preserves the 2D structural information of time and subspace in product-quantized latent code. Last, we employ a secondary stage to refine the preliminary predicted motion, further sharpening the high-frequency details, especially in facial movements.

To the best of our knowledge, our proposed method is the first approach to explicitly address the issue of holistic body variability and coordination in co-speech motion generation. 
We quantitatively evaluate the realism and diversity of our synthesized motion compared to ground truth and SOTA methods and ablations on the SHOW dataset \cite{yi2023generating}. Experimental results demonstrate that our model surpasses state-of-the-art methods both in qualitative and quantitative terms, with a particularly notable advancement in terms of realism.
\section{Related Work}

\subsection{Human Motion Generation}

Human motion generation is a popular research field and includes many sub-tasks. According to the input conditions, existing works in this field can be categorized into text-, audio-, and scene-driven motion generations \cite{zhu2023human}. 
The text-driven motion generation aims to generate human motion based on the action category \cite{guo2020action2motion, petrovich2021action, mao2022weakly}, or from textual descriptions \cite{plappert2016kit, guo2022generating, petrovich2022temos, zhang2023t2m, ma2024richcat}. 
The audio-driven motion generation focuses on generating conversational gestures in response to speech input, so-called co-speech motion generation \cite{ginosar2019gestures, yoon2020speech, yi2023generating}, or generating 3D dance motion based on audio input \cite{li2020learning, siyao2022bailando, zhuang2022music2dance}. 
The scene-driven motion generation is conditioned on the scene context \cite{wang2021synthesizing, araujo2023circle, ma2023grammar}.
Besides, some works \cite{harvey2020robust, cai2021unified, qin2022motion} target the motion completion task that utilizes sparse keyframes as constraints to generate a full motion sequence. 
We refer the readers to a recent review paper \cite{zhu2023human} for a comprehensive overview.

\subsection{Co-Speech Motion Generation}

Early approaches to co-speech motion generation are rule-based \cite{cassell2001beat, kopp2004synthesizing, levine2010gesture, poggi2005greta}. They adopt linguistic rules to translate speech into a sequence of pre-defined gesture segments. However, this process is time-consuming and labor-intensive as it involves expert knowledge and extensive effort to define the rules and segment the gestures. 
Recent attention has shifted towards learning-based methods, leveraging advancements in deep learning \cite{van2017neural, goodfellow2020generative, ho2020denoising}. These methods aim to estimate the mapping or probability distribution of co-speech motion, taking into account various condition modalities, including acoustic features \cite{ferstl2018investigating, kucherenko2019analyzing, li2021audio2gestures, zhu2023taming}, linguistic features \cite{ferstl2018investigating, kucherenko2019analyzing, li2021audio2gestures, zhu2023taming}, speaker identities \cite{yoon2020speech, liu2022learning}, and emotional cues \cite{qi2023emotiongesture, yin2023emog}.
Beyond exploring diverse co-speech motion conditions, many approaches employ various probabilistic models to generate a wide spectrum of co-speech motions, including Generative Adversarial Networks (GANs) \cite{ahuja2020style, yoon2020speech}, VAEs \cite{li2021audio2gestures}, VQ-VAE \cite{ao2022rhythmic, yazdian2022gesture2vec}, normalizing flows \cite{alexanderson2020style}, and diffusion models \cite{ao2023gesturediffuclip, zhu2023taming}.

\definecolor{newred}{RGB}{255, 0, 0}
\definecolor{newblue}{RGB}{65, 113, 156}
\definecolor{newgreen}{RGB}{169, 209, 142}
\newcommand{\frameworkCaption}{Overview of the proposed ProbTalk, a unified probabilistic framework designed to jointly model facial, hand, and body movements in speech. Specifically, ProbTalk first learns a PQ-VAE of holistic body motion, where the latent space is partitioned into subspaces, each of which is quantized using a dedicated codebook. Then, we predict the PQ codes based on the masked code context. In each iteration, we add 2D positional encoding (2D-PE) to embeddings of masked PQ codes to conserve the original structural integrity of the PQ codes. The predicted motion is refined in the secondary stage, further sharpening the high-frequency details.}

\begin{figure*}
    \centering
    \includegraphics[width=\textwidth]{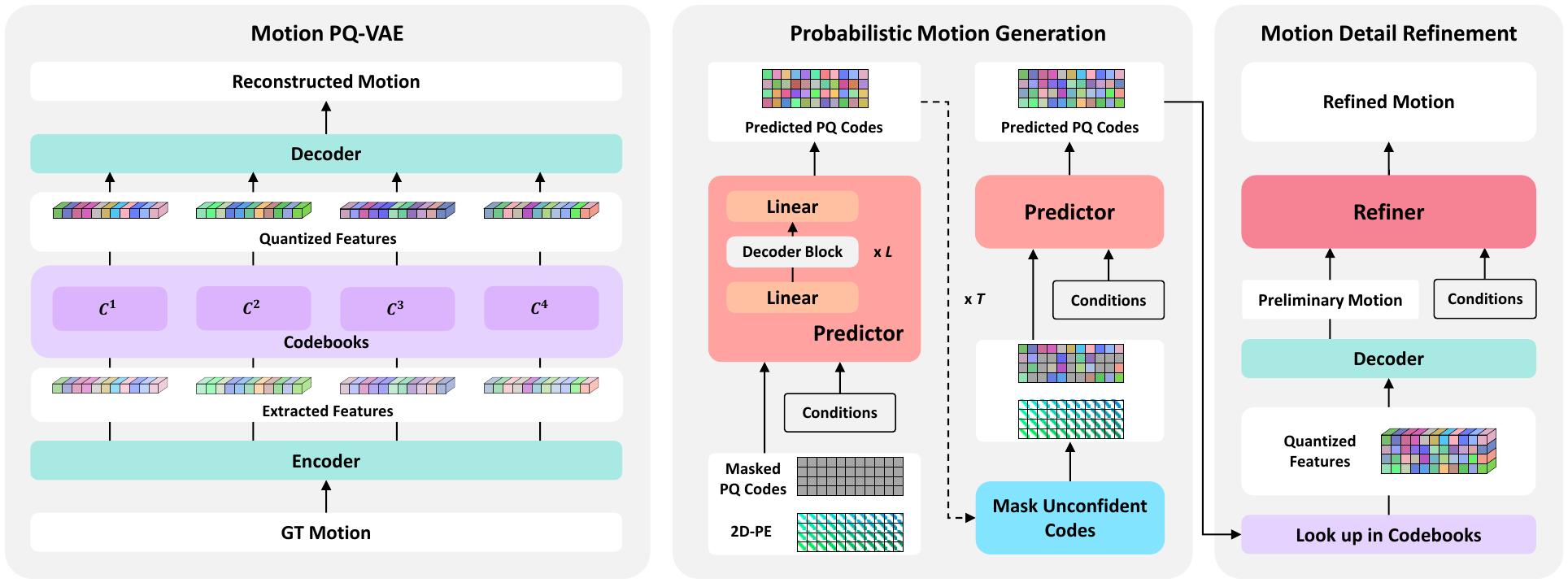}
    
    \caption{\frameworkCaption}
    \vspace{-0.04in}
    \label{fig:framework}
\end{figure*}

However, these methods primarily focus on body motion while overlooking face motion.
To address this limitation, TalkShow \cite{yi2023generating} introduced a separate modeling approach that creates probabilistic body movements while maintaining deterministic facial movements, enriching motion diversity and retaining facial detail. However, this method risks a lack of whole-body coordination, as it generates facial and body movements independently, potentially leading to unnatural alignment. To address this, Liu et al. \cite{liu2022beat} introduced a cascaded network architecture for synthesizing speech gestures based on ground truth (GT) facial movements. However, this approach is heavily dependent on the availability of GT facial data, which is generally not accessible in many scenarios. In a different vein, Habibie et al. \cite{habibie2021learning} proposed a method for the simultaneous generation of 3D whole-body motions associated with speech, thereby ensuring bodily coordination without the necessity for GT data. Despite their merits, both methods operate in a deterministic manner, limiting their variability, such as producing a range of diverse motions given the same speech input. 

Unlike previous method, we propose a unified probabilistic framework for co-speech motion generation. Our approach not only attains coordination between the facial and body movements but also ensures their motions are variable and diverse.

\section{Method} 

\paragraph{Preliminary.} 
We define a sequence of GT holistic body motion as $M_{1:N}=\{m_n|m_n\in \mathbb{R}^{376}\}^N_{n=1}$, where $N$ is the sequence length and each individual pose $m_n$ is represented by 6D \cite{zhou2019continuity} SMPL-X \cite{pavlakos2019expressive} parameters, including jaw pose, body pose, hand pose, and facial expression parameters.
The corresponding speech audio is encoded using a wav2vec 2.0 encoder \cite{baevski2020wav2vec}, leading to a sequence of audio features $A_{1:N}=\{a_n|a_n\in \mathbb{R}^{768}\}^N_{n=1}$. 

\subsection{Method Overview}
Given a speech recording, our goal is to synthesize variable and coordinated holistic body motion. To this end, we propose a unified probabilistic framework named \modelname to jointly model facial, hand, and body movements in speech. 
Compared to the deterministic method used by \cite{habibie2021learning}, our probabilistic framework facilitates the generation of more varied holistic co-speech motions.
It is also important to note that, unlike the previous method \cite{yi2023generating}, which models the face, body, and hands separately, our design has a noteworthy benefit in that facial expressions, hand gestures, and body poses are synthesized jointly. This crucial aspect of our approach ensures that the generated motions are not disconnected movements of individual body parts, but rather coordinated movements of the whole body. 

\subsection{Unified Probabilistic Motion Generation}
\label{sec: stage one}
To estimate diverse holistic human motion, we leverage the recent advances of PQ-VAE \cite{gray1984vector, wu2019learning} to learn a multi-mode distribution space for holistic body motions, thereby creating an expressive motion representation that allows us to capture the possible variability inherent in the holistic body motion. Concretely, we begin by encoding and quantizing the holistic body motion into $G$ groups of finite codebooks, from which we can sample a wide range of plausible body gestures (\cref{sec:pq}). Subsequently, we introduce a novel non-autoregressive model over the learned codebooks, enabling us to predict diverse body motions (\cref{sec:predictor}). Our Predictor is designed for efficient inference and effective prediction. Then, we obtain the preliminary holistic human motion by decoding codebook indices sampled from the distribution. Lastly, we refine the preliminary motion by a sequence-to-sequence model to capture intricate details, thus enabling a more accurate synchronization with the audio (\cref{sec: refiner}). \cref{fig:framework} illustrates our idea.

\subsubsection{Motion PQ-VAE}\setlength{\abovecaptionskip}{0pt}
\label{sec:pq}
Given a holistic motion sequence $M_{1:N}$, we aim to reconstruct the motion sequence by learning a discrete representation that can capture the intricate variations of holistic motion. 
The classical VQ-VAE, as described in \cite{van2017neural}, comprises an auto-encoder and a discrete codebook that learns to capture data patterns in the latent space of the auto-encoder. Due to the limited codebook combinations, VQ-VAE's ability to represent diverse data patterns is constrained by substantial quantization errors.
Drawing inspiration from prior research \cite{gray1984vector, wu2019learning}, we introduce a product quantization (PQ) approach to address this challenge. Specifically, PQ partitions the latent space into $G$ separate subspaces, each of which is represented by a separate sub-codebook. We denote the $g$-th sub-codebook as $C^g=\{c^g_k\}^K_{k=1}$. Similarly, the latent feature $z_n$ is partitioned into $G$ sub-features $\{z^{g}_n\}^G_{g=1}$ and then independently quantized with their corresponding codebook:
\begin{equation}
    \widehat{z}^g_n = \mathop{\arg\min_{c^g_k \in C^g}}\| c^g_k - z^g_n\| \in \mathbb{R}^{d^c},
\end{equation}
where $c^g_k$ denotes the embedding of the $k$-th code in the codebook $C^g$, and $d^c$ indicates the dimensionality of each code embedding.
Then, the quantized sub-features $\{\widehat{z}^g_n\}^G_{g=1}$
are fed into the decoder for the synthesis. Compared to classic vector quantization, PQ has an expansive representation capacity. For example, assuming that VQ and PQ have the same memory cost, the size of their respective code combinations are $K \times G$ and $K^G$, where $K \gg G$ (in our case, $K=128, G=4$). This means that PQ can handle exponentially larger data patterns with the same memory cost and reduce quantization errors. By leveraging PQ within the VAE, we can effectively capture the complex diversities of holistic motion.


The loss function $\mathcal{L}_{pq}$ for optimizing PQ-VAE consists of three components: a reconstruction loss, a velocity loss, and a ``commitment loss'' that encourages the feature to stay close to the chosen code. Let $V_{1:N-1} = \{v_n\}^{N-1}_{n=1}, v_n=m_n-m_{n+1}$ denotes the velocity of GT motions $M_{1:N}$, and $V^{pq}_{1:N-1} = \{v^{pq}_n\}^{N-1}_{n=1}, v^{pq}_n=m^{pq}_n-m^{pq}_{n+1}$ denotes the velocity of predicted motions $M^{pq}_{1:N}=\{m^{pq}_n\}^{N}_{n=1}$, the loss function $\mathcal{L}_{pq}$ can be formulated as:
\begin{equation}
\begin{aligned}
\mathcal{L}_{pq} = & \mathcal{L}_{1}(M_{1:N}, M^{pq}_{1:N})+\mathcal{L}_{1}(V_{1:N-1}, V^{pq}_{1:N-1}) \\
& + \beta\left\|Z-\operatorname{sg}\left[C\right]\right\|,
\end{aligned}
\end{equation}
where $\mathcal{L}_{1}$ is L1 reconstruction loss, $sg\left[\cdot\right]$ indicates stop gradient, and $\beta$ is the weight of the ``commitment loss''. 
Exponential moving average (EMA) and codebook reset (Code Reset) are used in the codebook updating \cite{razavi2019generating, zhang2023t2m}. 

\subsubsection{Non-autoregressive Modeling for Prediction}
\label{sec:predictor}
While applying PQ-VAE significantly enhances the ability to represent complex holistic motions with lower quantization errors, it also introduces specific challenges.
In particular, the longer sequence of latent code introduced by PQ-VAE requires more inference steps, thereby reducing the efficiency. Moreover, modeling the intricate relationships across the temporal sequence and subspace is not straightforward.

To tackle these issues, we propose a non-autoregressive model featuring two critical designs: MaskGIT and 2D positional encoding. Coupling these two designs enables us to train our Predictor efficiently and effectively, leading to 8x inference speedup and improved prediction performance.

\paragraph{MaskGIT-like Modeling.}
Due to the sequential nature of traditional autoregressive models, the increased code sequence length of PQ-VAE results in an augmented number of inference steps, reducing inference efficiency. Motivated by \cite{chang2022maskgit}, we tackle this issue by designing a MaskGIT-like Predictor that begins with generating all codes simultaneously and then refines the codes iteratively conditioned on the previous generation at inference time, as shown in \cref{fig:framework}.

In each training iteration, we initially get a sequence of pseudo-GT codes $X$ from PQ-VAE, represented as one-hot vectors. Subsequently, a binary mask $I$ is sampled, in which a value of 0 indicates that the corresponding elements in $X$ should be substituted with a distinct [MASK] code. This process is represented by the formula $\tilde{X} = I \odot X + (1-I) \odot X_{[MASK]}$, where $X_{[MASK]}$ is the index vector of [MASK]. The Predictor is trained to predict missing codes based on the conditions and unmasked codes.
Subsequently, the Predictor generates the logits of the codes $\widehat{X}$, taking into account multi-modal inputs, including audio $A$, 
motion context $M^c$, and identity $D$, as well as the masked code sequence $\tilde{X}$:
\begin{equation}
    \widehat{X} = \text{Predictor}(\tilde{X}; A, M^c, D).
\end{equation}

Then, we use cross entropy loss $\mathcal{L}_{ce}$ to optimize our Predictor:
\begin{equation}
    \mathcal{L}_{predict} = \mathcal{L}_{ce}(X, \widehat{X}).
\end{equation}

During inference, our model predicts all codes 
in parallel during each iteration. However, it only retains the most confident predictions and masks out the remaining codes for re-prediction in the next iteration. This iterative process continues until all codes are generated accurately.

\paragraph{2D Positional Encoding.}
Transformer models \cite{vaswani2017attention, devlin2018bert} conventionally represent both input and output data as one-dimensional sequences, incorporating a singular dimensional positional encoding to indicate the exact position of each element. However, this approach often overlooks certain explicit relational dynamics inherent to multi-dimensional data configurations. In our specific context, the intricate 2D structural correlations, which span both temporal sequences and various subspaces, may become obscured or less accessible when the PQ codes are linearized into a one-dimensional continuum. Consequently, converting PQ codes into a one-dimensional format requires the model to undergo a complex relearning process to decode and understand these interrelationships anew.

To circumvent this loss of multi-dimensional relational clarity, we design a bi-dimensional positional encoding strategy to conserve the original structural integrity of the PQ codes. This dual encoding schema assists the model in retaining and comprehending the complex interplay of relationships that exist between the codes, drawing inspiration from \cite{bello2019attention}. For the $n$-th code of the $g$-th subspace, the two-dimensional positional encoding ${e}_n^g$ is derived by summing the temporal encoding $\alpha_n$ and the spatial encoding $\beta_g$:
\begin{equation}
{e}_n^g = \alpha_n + \beta_g.
\end{equation}
Both $\alpha_n$ and $\beta_g$ are instantiated through sine and cosine functions \cite{vaswani2017attention}. This synthesis of temporal and subspace encodings seeks to retain a robust representation of the data’s multi-dimensional character within the Transformer model’s operational framework.

\newcommand{\layerCaption}{
Our framework is designed to generate co-speech motion, utilizing multi-modal conditions. In detail, the audio and motion context are individually processed by their respective condition encoders. Following this, the encoded outputs are concatenated and forwarded to a cross-attention layer. Besides, an AdaIN layer \cite{huang2017arbitrary} is integrated to facilitate the incorporation of speaker identity.
}
\begin{figure}
    \centering
    \includegraphics[width=\columnwidth]{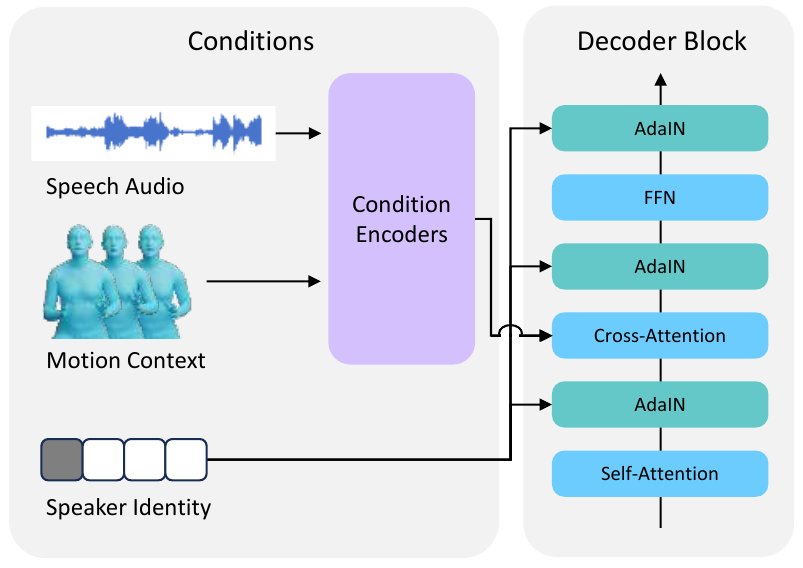}
    \caption{\layerCaption}
    \label{fig:layer}
\end{figure}

\subsubsection{Motion Detail Refinement} 
\label{sec: refiner}
The initially estimated motion outlines the overall style of the motion towards variability and coordination, yet neglects to capture the high-frequency details, particularly the swift transitions in facial movements. To further improve the preliminary motion, we introduce a motion detail refinement module to refine the preliminary motion utilizing a deterministic sequence-to-sequence model. This approach effectively captures intricate and fast-moving motion details, thereby enhancing the accuracy of the preliminary motion and achieving more precise synchronization with the audio. 

The Refiner has a similar network structure to the Predictor with the absence of MaskGIT-like modeling and 2D positional encoding. The Refiner takes combined motion $\tilde{M}_{1:N}$, input audio $A_{1:N}$, mask $I$ and speaker identity $D_{1:N}$ as input, and outputs refined motion $M^r_{1:N}$:
\begin{equation}
    M^r_{1:N} = \text{Refiner}(\tilde{M}_{1:N}; A_{1:N}, I, D).
\end{equation}

The combined motion $\tilde{M}_{1:N}$ is the combination of motion context $M^c_{1:N}$ and the preliminary motion $M^p_{1:N}$:
\begin{equation}
    \tilde{M}_{1:N} = I \odot M^c_{1:N} + \left( {1} - I\right) \odot M^p_{1:N}.
\end{equation}
Here, $M^c_{1:N}$ is generated by adding zeros to the desired motion prefix or suffix to fit our model's needs—this is known as ``zero padding''. On the other hand, $M^p_{1:N}$ comes from the output of a PQ-VAE's decoder, which takes predicted PQ codes as its input. 

More details are given in the supplemental material.

\definecolor{newyellow}{RGB}{255, 137, 0.}
\definecolor{newgrey}{rgb}{0.659, 0.659, 0.659}

\newcommand{\maskgitCaption}{
Qualitative comparison with SOTA methods. The co-speech motion generated by ProbTalk is more realistic, especially in terms of the timing, magnitude, and frequency of movements. We highlight the arm movements in \textcolor{newgrey}{grey}. Best viewed in color.
}

\begin{figure*}
    \centering
    \includegraphics[width=\linewidth]{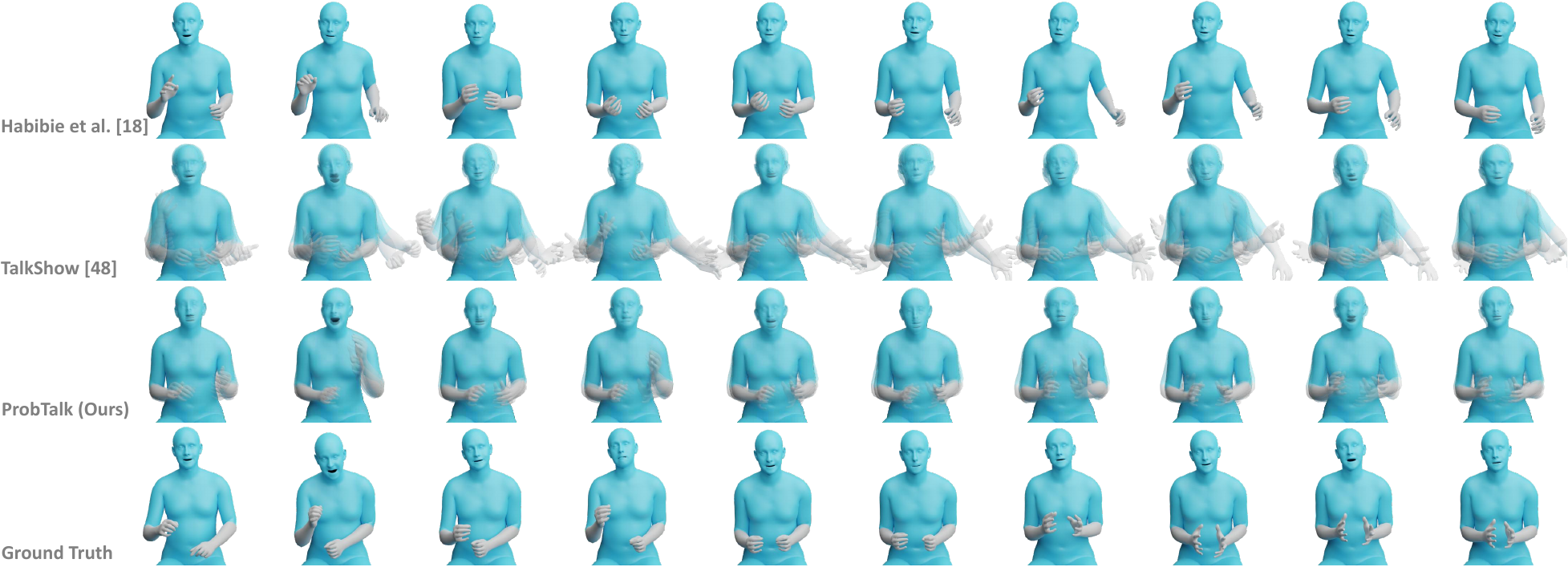}
    \caption{\maskgitCaption}
    \label{fig:qualitative results}
\end{figure*}

\subsection{Multi-Modal Conditioning} 
\label{multimodal-condition}
Finally, our framework is designed to support multi-modal conditioning. We incorporate modalities beyond audio, such as motion contexts, into our Predictor and Refiner models, as shown in \cref{fig:layer}. This enables us to facilitate motion completion and editing. Additionally, since speakers often present different motion styles, we utilize the modality of speaker identity to differentiate these styles.

\paragraph{Motion Contexts.} Our framework supports motion context as a speech modality, allowing for smooth and contextual-aware transitions in extended motion sequences. This leads to the capability of audio-driven motion completion. During training, the motion context is represented by the random masked ground truth (GT) motions, denoted as $M^c_{1:N} = I \odot M$. Here, $I \in \mathbb{R}^N$ represents the mask, and the unmasked frames serve as context frames. With the motion contexts, our Predictor generates preliminary motion by filling in the masked frames in a bidirectional manner. This preliminary motion is then refined by our Refiner, ensuring a seamless transition with the unmasked frames.

\paragraph{Speaker Identity.} Considering that speakers often display different motion styles, we leverage the modality of speaker identity $D$ to differentiate these styles, preventing shifts in motion style. Inspired by \cite{zhang2022motiondiffuse}, 
an Adaptive Instance Normalization (AdaIN) layer \cite{huang2017arbitrary} is introduced after each layer, facilitating the incorporation of speaker identity.

\section{Experiments}

\newcommand{\cmk}{\checkmark}

\begin{table}[t]
    \footnotesize
    \centering
    \resizebox{\linewidth}{!}{
    \begin{tabular}{cccccccccc}
    \toprule
 \multicolumn{4}{c}{Components} & \multicolumn{3}{c}{FGD $\downarrow$ }&\multicolumn{2}{c}{Var $\uparrow$} \\ \cmidrule(r){1-4} \cmidrule(r){5-7} \cmidrule(r){8-9}
    PQ& Mask &2D &R &  Holistic&  Body& Face& Body& Face & FPS $\uparrow$  \\\hline

     - & - & - & -                         & 6.51 & 7.32 & 26.06 & \textbf{0.47}  & 0.11 & 530          \\ 
    \checkmark & - & - & -                             & 4.41 & \textbf{5.17} & 15.69 & 0.39  & 0.16 & 132          \\
    \checkmark & \checkmark & - & -                    & 4.89 & 6.33 & 14.30 & 0.24  & 0.15 & \textbf{1176}          \\      
    \checkmark & \checkmark & \checkmark  & -          & 4.76 & 5.42 & 15.49 & 0.27  & 0.16 & 1071          \\
    - & \cmk & -  & \cmk                               & 6.93 & 9.22 & 8.59 & 0.26  & 0.17 & 1048 \\
    \cmk & - & -  & \cmk                               & 4.27 & 5.37 & 6.38 & 0.38  & \textbf{0.20} & 132 \\
    \cmk & \cmk & -  & \cmk                            & 4.65 & 6.11 & 5.77 & 0.23  & 0.16 & 1039        \\
    \cmk & \cmk & \cmk  & \cmk                    & \textbf{3.98} & 5.21 & \textbf{5.59} & 0.26  & 0.17 & 1067  \\\hline
    \end{tabular}
    }
    \caption{Ablation study on each key component of our methods. ``Mask'' denotes the MaskGIT-like modeling, ``2D'' denotes the 2D postional encoding, and ``R'' denotes the Refiner.}
    \label{tab:ablation}
\end{table}

\begin{table}[t]
    \footnotesize
    \centering
    \resizebox{\columnwidth}{!}{
    \begin{tabular}{lcccccc}
    \toprule
    & \multicolumn{3}{c}{FGD $\downarrow$ }&\multicolumn{2}{c}{Var $\uparrow$} \\ \cmidrule(r){2-4} \cmidrule(r){5-6}
    & \tabincell{c}{Holistic}
    & \tabincell{c}{Body}
    & \tabincell{c}{Face}
    & \tabincell{c}{Body}
    & \tabincell{c}{Face}
    & \tabincell{c}{FPS $\uparrow$} \\ \hline
    Habibie et al. \cite{habibie2021learning}   & 44.60 & 44.17 & 43.66 & 0. & 0. & \textbf{24733} \\
    TalkSHOW \cite{yi2023generating}       & 6.60 & 7.93 & 7.69 & \textbf{0.94} & 0. & 205 \\ \hline
    Separate        & 4.47 & 5.32 & 6.63 & 0.20 & 0. & 991 \\
    Simultaneous-R    & 7.53 & 11.11 & 6.59 & 0. & 0. & 8990 \\
    Simultaneous-P    & 4.76 & 5.42  & 15.49 & 0.27 & 0.16 & 1071 \\
    ProbTalk (Ours)    & \textbf{3.98} & \textbf{5.21}  & \textbf{5.59} & 0.26 & \textbf{0.17} & 1067 \\\hline
    \end{tabular}
    }
    \caption{Comparison with state-of-the-arts and various baselines.}
    \label{tab:structure}
\end{table}
\subsection{Dataset}

In this section, we evaluate our approach on the SHOW dataset \cite{yi2023generating}. It is a database of 3D holistic body mesh annotations with synchronous audio from in-the-wild videos, including 26.9 hours of talkshow videos from 4 speakers. 
We select video sequences longer than 6 seconds and divide the dataset into 80-10-10\% train, validation and unseen test splits.

\subsection{Experimental Setup}

\paragraph{Implementation Details.}
Our framework is trained sequentially, beginning with the PQ-VAE, followed by the Predictor, and lastly, the Refiner. We employ AdamW as the optimizer, with $[\beta_1, \beta_2] = [0.9, 0.99]$ and a learning rate of 0.0001 for each of the three parts. A batch size of 128 is used to train all three parts for 100 epochs. The window size $\tau$ is set to 8. 

For the PQ-VAE, the number of codebooks is $G=4$, and for each codebook, the number of codes is $K=128$. The weight of the ``commitment loss'' term is set to $\beta=0.25$. For the Predicter, the mask ratio is controlled via a cosine scheduler, following \cite{chang2022maskgit}, and the number of iterations during inference is $T=8$.

\paragraph{Evaluation Metrics.} We evaluate the performance of various approaches from multiple perspectives, which encompass realism, diversity, and inference efficiency. Specifically, we employ the following commonly used metrics:
\begin{itemize}[leftmargin=*]
\item \ \textit{FGD}: \textbf{F}r\'echet \textbf{G}esture \textbf{D}istance is proposed by Yoon et al. \cite{yoon2020speech}, which measures the difference between the distributions of the latent features of the generated gestures and ground truth. We report three variants of FGD: Holistic, Body, and Face. These metrics serve to evaluate the perceived plausibility of the respective synthesized gesture components.
\item \  \textit{Variance}: As used in \cite{ng2022learning, yi2023generating}, diversity is assessed by quantifying the variance among multiple samples originating from the same condition.
\item \  \textit{FPS}: \textbf{F}rame \textbf{P}er \textbf{S}econd refers to the number of individual frames that are processed in one second, measuring the inference efficiency. In our experiment, we evaluate the FPS performance of each method using a TITAN Xp GPU with a batch size of 1.
\end{itemize}

Besides, we report both the first-order and second-order L2 errors
to measure the reconstruction performance of PQ-VAE.

\subsection{Qualitative Analysis}
In our qualitative comparison of ProbTalk with TalkShow \cite{yi2023generating} and the method by Habibie et al. \cite{habibie2021learning}, we aim to showcase the superior generation quality of our approach. Utilizing the same speech recording, we generated nine samples with each method. These samples are overlaid for a direct comparison with the Ground Truth (GT), as depicted in \cref{fig:qualitative results}. The figure clearly demonstrates that the GT motion includes four instances of swinging movements within a given time frame. The methods of Habibie et al. and TalkShow fail to capture this repetitive motion pattern, leading to motions that are slower and less lifelike. In contrast, ProbTalk accurately captures the timing, magnitude, and frequency of these swings, showcasing its precision in emulating realistic motion dynamics.

\subsection{Quantitative Analysis}

\paragraph{Comparison with State-of-the-Art Methods.} 
In our comparative analysis, presented in Table \ref{tab:structure}, our method demonstrates superior performance over existing state-of-the-art methods across all FGD metrics. Notably, our approach shows significant advancements in FGD (Holistic), indicating its capacity for generating realistic holistic body motions. This can be attributed to our proposed unified probabilistic framework. Additionally, our method excels in Var (face), indicating its effectiveness in producing diverse facial expressions. Compared to the TalkShow model \cite{yi2023generating}, our approach also achieves a higher FPS, highlighting both its efficiency and effectiveness.

To further showcase the effectiveness of our framework design, we evaluate various baseline models, ensuring a fair comparison by employing similar techniques such as PQ-VAE, MaskGIT, and 2D positional encoding. The results are presented in \cref{tab:structure}. The ``Separate'' model, akin to TalkShow \cite{yi2023generating}, treats body and facial motions independently. Conversely, the ``Simultaneous'' model, following Habibie et al. \cite{habibie2021learning}, where the entire body motion is modeled jointly, employing either a ``P''redictor or a ``R''efiner. Our results indicate that ``Simultaneous-R'' performs the least effectively in terms of FGD (Holistic) and FGD (Body), underscoring the strength of our probabilistic approach in body motion generation. Our method shows significant improvements in FGD (Holistic) and FGD (Face) compared to ``Simultaneous-P'', emphasizing the importance of the Refiner in our framework. Notably, our method surpasses the ``Separate'' model, demonstrating the benefits of joint modeling. 

\paragraph{Model Ablation.} 
We evaluate the effect of each component and report the results 
in \cref{tab:ablation}, from which several important conclusions are drawn:

\textbf{$\bigcdot$} \ PQ plays a crucial role in enhancing the realism of the generated motion. Nevertheless, it comes with a trade-off in terms of inference efficiency. When compared to the baseline, incorporating PQ leads to significant improvements in FGD. However, it also results in inefficiency during inference due to the lengthier code sequence, which requires more iterations in the autoregressive inference process.

\textbf{$\bigcdot$} \ The integration of MaskGIT proves to be effective in achieving faster inference times, as evidenced by a significant boost in FPS. However, this improvement comes at the expense of a decrease in the realism of the generated motion. While MaskGIT enhances the efficiency of our method, it also negatively impacts the FGD (Body), which in turn affects the overall realism of the holistic body motion, as indicated by the FGD (Holistic) evaluation.

\textbf{$\bigcdot$} \ 2D positional encoding (2D-PE) proves to be beneficial in generating more realistic holistic body motion. When compared to the method without 2D-PE, incorporating 2D-PE leads to significant improvements in various FGD metrics. This highlights the importance and effectiveness of 2D-PE in enhancing the quality and realism of the generated motion, especially when the integration of MaskGIT results in a decrease in the realism of the generated motion.

\textbf{$\bigcdot$} \ Refiner proves beneficial in refining high-frequency details, particularly in the case of face motion. Incorporating this design yields significant improvements in FGD (face). Furthermore, it also contributes to the overall generation of more lifelike body motion, as demonstrated by the observed improvements in FGD (body).

\subsection{Sensitive Analysis}

\begin{table}[t]
    \centering
    \footnotesize
    \resizebox{\columnwidth}{!}{
    \begin{tabular}{llccccc}
    \toprule

    & & \multicolumn{3}{c}{FGD $\downarrow$ }&\multicolumn{2}{c}{L2 Errors $\downarrow$} \\ \cmidrule(r){3-5} \cmidrule(r){6-7}
    $K$ & $G$
    & \tabincell{c}{Holistic}
    & \tabincell{c}{Body}
    & \tabincell{c}{Face}
    & \tabincell{c}{1st-order}
    & \tabincell{c}{2nd-order} 
    
    \\\hline
    512 & 1     &5.43&6.11&24.91 &	3.47 	&0.21 \\
    1024 & 1    &4.73&5.37&22.49 &	3.41 	&0.20  \\
    2048 & 1    &4.73&5.49&22.29 &	3.38 	&0.20 \\
    4096 & 1    &5.05&5.98&24.57 &	3.35 	&0.20 \\
    128 & 2     &3.60&4.30&19.36 &	3.10 	&0.20 \\ 
    128 & 4     &\textbf{1.72}&\textbf{1.88}&\textbf{13.10} &	\textbf{2.71} 	&\textbf{0.18} \\
    \hline
    \end{tabular}
    }
    \caption{Reconstruction performance of PQ-VAE with different codebook size $K$ and the number of codebooks $G$.}
    \label{tab:effect_of_product}

\end{table}

\paragraph{Effect of Product Quantization.}  
To demonstrate the superior representation capability of PQ, we present the reconstruction performance of VAE across various codebook sizes $K$ and the number of codebooks $G$. The results are summarized in \cref{tab:effect_of_product}. It is noteworthy that increasing the codebook size $K$ does not yield a significant improvement in the model's performance concerning MAJE and MAD. Moreover, it adversely affects the model's performance in terms of FGD. This observation suggests that a relatively small codebook size $K$ suffices, primarily due to the implementation of advanced codebook update strategies, namely, EMA and CodeReset. Conversely, a substantial improvement is evident across almost all metrics as we increase the number of codebooks $G$, highlighting PQ-VAE's proficiency in representing realistic human motion.

\begin{table}[t]
    \centering
    \resizebox{\linewidth}{!}{
    \begin{tabular}{cccccccc}
    \toprule
    & & \multicolumn{3}{c}{FGD $\downarrow$ }&\multicolumn{2}{c}{Var $\uparrow$} \\ \cmidrule(r){3-5} \cmidrule(r){6-7}
    & $T$
    & \tabincell{c}{Holistic}
    & \tabincell{c}{Body}
    & \tabincell{c}{Face}
    & \tabincell{c}{Body}
    & \tabincell{c}{Face}
    & \tabincell{c}{FPS $\uparrow$} \\ \hline
    Autoregressive & -       & 4.27 & 5.37 & 6.38 & \textbf{0.38} & \textbf{0.20} & 132\\ \hline
    \multirow{5}{*}{MaskGIT}
    & 1                      & 4.99 & 6.95 & 5.74 & 0.30 & 0.19 & \textbf{3660} \\
    & 2                      & 4.29 & 5.82 & 5.59 & 0.28 & 0.18 & 2689 \\ 
    & 4                      & 4.03 & 5.34 & 5.60 & 0.27 & 0.17 & 1770 \\
    & 8                      & 3.98 & 5.21 & 5.59 & 0.26 & 0.17 & 1067 \\ 
    & 16                     & \textbf{3.97} & \textbf{5.16} & \textbf{5.56} & 0.25 & 0.17 & 569 \\ \hline
    \end{tabular}
    }
    \caption{Performance of MaskGIT with different number of iterations $T$.}
    \label{tab:effect_of_MaskGIT}

\end{table}

\paragraph{Effect of MaskGIT.} 
To evaluate the impact of MaskGIT, we conducted a comprehensive analysis of the model's performance across varying numbers of iterations, denoted as $T$. The results are presented in \cref{tab:effect_of_MaskGIT}. In comparison to autoregressive modeling, MaskGIT demonstrates a notable enhancement in inference efficiency, albeit accompanied by a reduction in the realism of the generated motion.
As we observe the influence of the number of iterations ($T$) on the performance metrics, an important trend emerges. An increase in $T$ leads to a significant rise in the FGD metrics, indicating improved motion quality. However, this improvement is countered by a slight decrease in variance and inference efficiency.
Furthermore, when we extend $T$ from 8 to 16, we do not observe a substantial increase in either the FGD score or the model's variance. 
This phenomenon suggests that there exists an optimal range of iteration counts within which realism enhancement plateaus.
In conclusion, the judicious selection of the iteration count parameter, $T$, allows for striking a balance between markedly improved inference efficiency and the preservation of motion realism.

\begin{table}[t]
    \footnotesize
    \centering
    \resizebox{\columnwidth}{!}{
    \begin{tabular}{lccccc}
    \toprule
    & \multicolumn{3}{c}{FGD $\downarrow$ }&\multicolumn{2}{c}{Var $\uparrow$} \\ \cmidrule(r){2-4} \cmidrule(r){5-6}
    & \tabincell{c}{Holistic}
    & \tabincell{c}{Body}
    & \tabincell{c}{Face}
    & \tabincell{c}{Body}
    & \tabincell{c}{Face} \\ \hline
    1D-Sine             & 4.65 & 6.11 & 5.77 & 0.23 & 0.16 \\
    1D-Trainable        & 4.51 & 6.07 & 5.72 & 0.25 & 0.17 \\
    2D-Sine             & \textbf{3.98} & \textbf{5.21} & \textbf{5.59} & \textbf{0.26} & \textbf{0.19} \\
    2D-Trainable        & 5.03 & 6.84 & 6.12 & 0.21 & 0.15 \\\hline
    \end{tabular}
    }
    \caption{Performance of different positional encoding methods.}
    \label{tab:pe}
\end{table}

\paragraph{Effect of the Positional Encoding.} 
We conducted an assessment of the impact of various positional encoding methods, and the results are presented in \cref{tab:pe}. ``Sine'' indicates that positional encoding parameters are generated using sine and cosine functions \cite{vaswani2017attention}, whereas ``Trainable'' indicates that these parameters are learned during the training process \cite{devlin2018bert}. The data in the table clearly illustrates that 2D positional encoding, formulated through sine and cosine functions, significantly outperforms other methods across all evaluated metrics. In contrast, 2D positional encoding with learned parameters exhibits the worst results. This disparity underscores the critical role of stable structural information in positional encoding.

\begin{table}[t]
    \footnotesize
    \centering
    \resizebox{\columnwidth}{!}{
    \begin{tabular}{lcccccc}
    \toprule
    & \multicolumn{3}{c}{FGD $\downarrow$ }&\multicolumn{2}{c}{Var $\uparrow$} \\ \cmidrule(r){2-4} \cmidrule(r){5-6}
    & \tabincell{c}{Holistic}
    & \tabincell{c}{Body}
    & \tabincell{c}{Face}
    & \tabincell{c}{Body}
    & \tabincell{c}{Face} \\ \hline
    Audio            & 6.18 & 6.99 & 9.65 & \textbf{0.33} & \textbf{0.21}  \\ 
    + Identity       & 4.75 & 6.49 & 5.61 & 0.24 & 0.19  \\ 
    + Motion Context & \textbf{3.98} & \textbf{5.21} & \textbf{5.59} & 0.26 & 0.19 \\ 
    \hline
    \end{tabular}
    }
    \caption{Effect of different input modalities.}
    \label{tab:effect_of_text}

\end{table}

\paragraph{Effect of Multi-Modal Conditioning.} 
To assess the effect of each input modality, we conduct a sequence of incremental experiments. The results are detailed in \cref{tab:effect_of_text}. The data presented in this table clearly indicates that each input modality significantly enhances the realism of the generated motion. Notably, the integration of identity information imposes a constraint on the motion style, thereby ensuring stylistic consistency throughout the motion sequence. Additionally, the inclusion of motion context plays a crucial role in facilitating the generation of smoother and more continuous movements, particularly when creating extended motion sequences.

\section{Conclusion}
In this study, we introduce ProbTalk, the first approach specifically designed to tackle the challenges of holistic body variability and coordination in the co-speech motion generation. The first step in our approach is the incorporation of PQ into the VAE, which significantly enhances the representation of complex, holistic motions. Next, we develop a unique non-autoregressive model that integrates 2D positional encoding, leading to efficient and effective inference. Finally, we utilize a secondary stage to refine the initial predictions, thereby improving the richness of high-frequency details. The experimental results validate that our approach delivers state-of-the-art performance in both qualitative and quantitative aspects.

\medskip

\noindent
{\qheading{Acknowledgments.}
We thank Hongwei Yi for the insightful discussions and Hualin Liang for helping us conduct the user study.
This work was partially supported by the Major Science and Technology Innovation 2030 ``New Generation Artificial Intelligence” key project (No. 2021ZD0111700), the National Natural Science Foundation of China under Grant 62076101, Guangdong Basic and Applied Basic Research Foundation under Grant 2023A1515010007, the Guangdong Provincial Key Laboratory of Human Digital Twin under Grant 2022B1212010004, and the TCL Young Scholars Program.
}

\bibliography{main}

\begin{thebibliography}{67}
\providecommand{\natexlab}[1]{#1}
\providecommand{\url}[1]{\texttt{#1}}
\expandafter\ifx\csname urlstyle\endcsname\relax
  \providecommand{\doi}[1]{doi: #1}\else
  \providecommand{\doi}{doi: \begingroup \urlstyle{rm}\Url}\fi

\bibitem[Ahuja et~al.(2020)Ahuja, Lee, Nakano, and Morency]{ahuja2020style}
Chaitanya Ahuja, Dong~Won Lee, Yukiko~I Nakano, and Louis-Philippe Morency.
\newblock Style transfer for co-speech gesture animation: A multi-speaker conditional-mixture approach.
\newblock In \emph{{European Conference on Computer Vision (ECCV)}}, pages 248--265. Springer, 2020.

\bibitem[Alexanderson et~al.(2020)Alexanderson, Henter, Kucherenko, and Beskow]{alexanderson2020style}
Simon Alexanderson, Gustav~Eje Henter, Taras Kucherenko, and Jonas Beskow.
\newblock Style-controllable speech-driven gesture synthesis using normalising flows.
\newblock In \emph{{Computer Graphics Forum (CGF)}}, pages 487--496. Wiley Online Library, 2020.

\bibitem[Ao et~al.(2022)Ao, Gao, Lou, Chen, and Liu]{ao2022rhythmic}
Tenglong Ao, Qingzhe Gao, Yuke Lou, Baoquan Chen, and Libin Liu.
\newblock Rhythmic gesticulator: Rhythm-aware co-speech gesture synthesis with hierarchical neural embeddings.
\newblock \emph{ACM Transactions on Graphics (TOG)}, 41\penalty0 (6):\penalty0 1--19, 2022.

\bibitem[Ao et~al.(2023)Ao, Zhang, and Liu]{ao2023gesturediffuclip}
Tenglong Ao, Zeyi Zhang, and Libin Liu.
\newblock Gesturediffuclip: Gesture diffusion model with clip latents.
\newblock \emph{ACM Transactions on Graphics (TOG)}, 42\penalty0 (4):\penalty0 1--18, 2023.

\bibitem[Ara{\'u}jo et~al.(2023)Ara{\'u}jo, Li, Vetrivel, Agarwal, Wu, Gopinath, Clegg, and Liu]{araujo2023circle}
Joao~Pedro Ara{\'u}jo, Jiaman Li, Karthik Vetrivel, Rishi Agarwal, Jiajun Wu, Deepak Gopinath, Alexander~William Clegg, and Karen Liu.
\newblock Circle: Capture in rich contextual environments.
\newblock In \emph{{Computer Vision and Pattern Recognition (CVPR)}}, pages 21211--21221, 2023.

\bibitem[Baevski et~al.(2020)Baevski, Zhou, Mohamed, and Auli]{baevski2020wav2vec}
Alexei Baevski, Yuhao Zhou, Abdelrahman Mohamed, and Michael Auli.
\newblock wav2vec 2.0: A framework for self-supervised learning of speech representations.
\newblock In \emph{{Conference on Neural Information Processing Systems (NeurIPS)}}, pages 12449--12460, 2020.

\bibitem[Bello et~al.(2019)Bello, Zoph, Vaswani, Shlens, and Le]{bello2019attention}
Irwan Bello, Barret Zoph, Ashish Vaswani, Jonathon Shlens, and Quoc~V Le.
\newblock Attention augmented convolutional networks.
\newblock In \emph{{International Conference on Computer Vision ({ICCV})}}, pages 3286--3295, 2019.

\bibitem[Cai et~al.(2021)Cai, Wang, Zhu, Cham, Cai, Yuan, Liu, Zheng, Yan, Ding, et~al.]{cai2021unified}
Yujun Cai, Yiwei Wang, Yiheng Zhu, Tat-Jen Cham, Jianfei Cai, Junsong Yuan, Jun Liu, Chuanxia Zheng, Sijie Yan, Henghui Ding, et~al.
\newblock A unified 3d human motion synthesis model via conditional variational auto-encoder.
\newblock In \emph{{International Conference on Computer Vision ({ICCV})}}, pages 11645--11655, 2021.

\bibitem[Cassell et~al.(2001)Cassell, Vilhj{\'a}lmsson, and Bickmore]{cassell2001beat}
Justine Cassell, Hannes~H{\"o}gni Vilhj{\'a}lmsson, and Timothy Bickmore.
\newblock Beat: the behavior expression animation toolkit.
\newblock In \emph{{International Conference on Computer Graphics and Interactive Techniques (SIGGRAPH)}}, pages 477--486, 2001.

\bibitem[Chang et~al.(2022)Chang, Zhang, Jiang, Liu, and Freeman]{chang2022maskgit}
Huiwen Chang, Han Zhang, Lu Jiang, Ce Liu, and William~T Freeman.
\newblock Maskgit: Masked generative image transformer.
\newblock In \emph{{Computer Vision and Pattern Recognition (CVPR)}}, pages 11315--11325, 2022.

\bibitem[Devlin et~al.(2018)Devlin, Chang, Lee, and Toutanova]{devlin2018bert}
Jacob Devlin, Ming-Wei Chang, Kenton Lee, and Kristina Toutanova.
\newblock Bert: Pre-training of deep bidirectional transformers for language understanding.
\newblock \emph{arXiv preprint arXiv:1810.04805}, 2018.

\bibitem[Ding and Tao(2018)]{ding2018trunk}
Changxing Ding and Dacheng Tao.
\newblock Trunk-branch ensemble convolutional neural networks for video-based face recognition.
\newblock \emph{{Transactions on Pattern Analysis and Machine Intelligence (TPAMI)}}, 40\penalty0 (04):\penalty0 1002--1014, 2018.

\bibitem[Fan et~al.(2022)Fan, Lin, Saito, Wang, and Komura]{fan2022faceformer}
Yingruo Fan, Zhaojiang Lin, Jun Saito, Wenping Wang, and Taku Komura.
\newblock {FaceFormer}: Speech-driven {3D} facial animation with transformers.
\newblock In \emph{{Computer Vision and Pattern Recognition (CVPR)}}, pages 18770--18780, 2022.

\bibitem[Ferstl and McDonnell(2018)]{ferstl2018investigating}
Ylva Ferstl and Rachel McDonnell.
\newblock Investigating the use of recurrent motion modelling for speech gesture generation.
\newblock In \emph{{International Conference on Intelligent Virtual Agents (IVA)}}, pages 93--98, 2018.

\bibitem[Ginosar et~al.(2019)Ginosar, Bar, Kohavi, Chan, Owens, and Malik]{ginosar2019gestures}
S. Ginosar, A. Bar, G. Kohavi, C. Chan, A. Owens, and J. Malik.
\newblock Learning individual styles of conversational gesture.
\newblock In \emph{{Computer Vision and Pattern Recognition (CVPR)}}, 2019.

\bibitem[Goldin-Meadow(1999)]{goldin1999role}
Susan Goldin-Meadow.
\newblock The role of gesture in communication and thinking.
\newblock \emph{Trends in Cognitive Sciences}, 3\penalty0 (11):\penalty0 419--429, 1999.

\bibitem[Goodfellow et~al.(2020)Goodfellow, Pouget-Abadie, Mirza, Xu, Warde-Farley, Ozair, Courville, and Bengio]{goodfellow2020generative}
Ian Goodfellow, Jean Pouget-Abadie, Mehdi Mirza, Bing Xu, David Warde-Farley, Sherjil Ozair, Aaron Courville, and Yoshua Bengio.
\newblock Generative adversarial networks.
\newblock \emph{Communications of the ACM}, 63\penalty0 (11):\penalty0 139--144, 2020.

\bibitem[Gray(1984)]{gray1984vector}
Robert Gray.
\newblock Vector quantization.
\newblock \emph{IEEE Assp Magazine}, 1\penalty0 (2):\penalty0 4--29, 1984.

\bibitem[Guo et~al.(2020)Guo, Zuo, Wang, Zou, Sun, Deng, Gong, and Cheng]{guo2020action2motion}
Chuan Guo, Xinxin Zuo, Sen Wang, Shihao Zou, Qingyao Sun, Annan Deng, Minglun Gong, and Li Cheng.
\newblock Action2motion: Conditioned generation of 3d human motions.
\newblock In \emph{Proceedings of the 28th ACM International Conference on Multimedia}, pages 2021--2029, 2020.

\bibitem[Guo et~al.(2022)Guo, Zou, Zuo, Wang, Ji, Li, and Cheng]{guo2022generating}
Chuan Guo, Shihao Zou, Xinxin Zuo, Sen Wang, Wei Ji, Xingyu Li, and Li Cheng.
\newblock Generating diverse and natural 3d human motions from text.
\newblock In \emph{{Computer Vision and Pattern Recognition (CVPR)}}, pages 5152--5161, 2022.

\bibitem[Habibie et~al.(2021)Habibie, Xu, Mehta, Liu, Seidel, Pons-Moll, Elgharib, and Theobalt]{habibie2021learning}
Ikhsanul Habibie, Weipeng Xu, Dushyant Mehta, Lingjie Liu, Hans-Peter Seidel, Gerard Pons-Moll, Mohamed Elgharib, and Christian Theobalt.
\newblock Learning speech-driven {3D} conversational gestures from video.
\newblock In \emph{{International Conference on Intelligent Virtual Agents (IVA)}}, pages 101--108, 2021.

\bibitem[Harvey et~al.(2020)Harvey, Yurick, Nowrouzezahrai, and Pal]{harvey2020robust}
F{\'e}lix~G Harvey, Mike Yurick, Derek Nowrouzezahrai, and Christopher Pal.
\newblock Robust motion in-betweening.
\newblock \emph{{Transactions on Graphics (TOG)}}, 39\penalty0 (4):\penalty0 60--1, 2020.

\bibitem[Ho et~al.(2020)Ho, Jain, and Abbeel]{ho2020denoising}
Jonathan Ho, Ajay Jain, and Pieter Abbeel.
\newblock Denoising diffusion probabilistic models.
\newblock \emph{Advances in Neural Information Processing Systems}, 33:\penalty0 6840--6851, 2020.

\bibitem[Huang and Belongie(2017)]{huang2017arbitrary}
Xun Huang and Serge Belongie.
\newblock Arbitrary style transfer in real-time with adaptive instance normalization.
\newblock In \emph{{International Conference on Computer Vision ({ICCV})}}, pages 1501--1510, 2017.

\bibitem[Ioffe and Szegedy(2015)]{ioffe2015batch}
Sergey Ioffe and Christian Szegedy.
\newblock Batch normalization: Accelerating deep network training by reducing internal covariate shift.
\newblock In \emph{{International Conference on Machine Learning (ICML)}}, pages 448--456, 2015.

\bibitem[Kendon(2004)]{kendon2004gesture}
Adam Kendon.
\newblock \emph{Gesture: Visible action as utterance}.
\newblock Cambridge University Press, 2004.

\bibitem[Kopp and Wachsmuth(2004)]{kopp2004synthesizing}
Stefan Kopp and Ipke Wachsmuth.
\newblock Synthesizing multimodal utterances for conversational agents.
\newblock \emph{Computer Animation and Virtual Worlds}, 15\penalty0 (1):\penalty0 39--52, 2004.

\bibitem[Kucherenko et~al.(2019)Kucherenko, Hasegawa, Henter, Kaneko, and Kjellstr{\"o}m]{kucherenko2019analyzing}
Taras Kucherenko, Dai Hasegawa, Gustav~Eje Henter, Naoshi Kaneko, and Hedvig Kjellstr{\"o}m.
\newblock Analyzing input and output representations for speech-driven gesture generation.
\newblock In \emph{{International Conference on Intelligent Virtual Agents (IVA)}}, pages 97--104, 2019.

\bibitem[Levine et~al.(2010)Levine, Kr{\"a}henb{\"u}hl, Thrun, and Koltun]{levine2010gesture}
Sergey Levine, Philipp Kr{\"a}henb{\"u}hl, Sebastian Thrun, and Vladlen Koltun.
\newblock Gesture controllers.
\newblock In \emph{{International Conference on Computer Graphics and Interactive Techniques (SIGGRAPH)}}, pages 1--11, 2010.

\bibitem[Li et~al.(2020)Li, Yin, Chu, Zhou, Wang, Fidler, and Li]{li2020learning}
Jiaman Li, Yihang Yin, Hang Chu, Yi Zhou, Tingwu Wang, Sanja Fidler, and Hao Li.
\newblock Learning to generate diverse dance motions with transformer.
\newblock \emph{arXiv preprint arXiv:2008.08171}, 2020.

\bibitem[Li et~al.(2021)Li, Kang, Pei, Zhe, Zhang, He, and Bao]{li2021audio2gestures}
Jing Li, Di Kang, Wenjie Pei, Xuefei Zhe, Ying Zhang, Zhenyu He, and Linchao Bao.
\newblock {Audio2Gestures}: Generating diverse gestures from speech audio with conditional variational autoencoders.
\newblock In \emph{{Computer Vision and Pattern Recognition (CVPR)}}, pages 11293--11302, 2021.

\bibitem[Liang et~al.(2022)Liang, Feng, Zhu, Hu, Pan, and Yang]{liang2022seeg}
Yuanzhi Liang, Qianyu Feng, Linchao Zhu, Li Hu, Pan Pan, and Yi Yang.
\newblock Seeg: Semantic energized co-speech gesture generation.
\newblock In \emph{{Computer Vision and Pattern Recognition (CVPR)}}, pages 10473--10482, 2022.

\bibitem[Liu et~al.(2022{\natexlab{a}})Liu, Iwamoto, Zhu, Li, Zhou, Bozkurt, and Zheng]{liu2022disco}
Haiyang Liu, Naoya Iwamoto, Zihao Zhu, Zhengqing Li, You Zhou, Elif Bozkurt, and Bo Zheng.
\newblock Disco: Disentangled implicit content and rhythm learning for diverse co-speech gestures synthesis.
\newblock In \emph{Proceedings of the 30th ACM International Conference on Multimedia}, pages 3764--3773, 2022{\natexlab{a}}.

\bibitem[Liu et~al.(2022{\natexlab{b}})Liu, Zhu, Iwamoto, Peng, Li, Zhou, Bozkurt, and Zheng]{liu2022beat}
Haiyang Liu, Zihao Zhu, Naoya Iwamoto, Yichen Peng, Zhengqing Li, You Zhou, Elif Bozkurt, and Bo Zheng.
\newblock Beat: A large-scale semantic and emotional multi-modal dataset for conversational gestures synthesis.
\newblock In \emph{Computer Vision--ECCV 2022: 17th European Conference, Tel Aviv, Israel, October 23--27, 2022, Proceedings, Part VII}, pages 612--630. Springer, 2022{\natexlab{b}}.

\bibitem[Liu et~al.(2023)Liu, Zhu, Becherini, Peng, Su, Zhou, Iwamoto, Zheng, and Black]{liu2023emage}
Haiyang Liu, Zihao Zhu, Giorgio Becherini, Yichen Peng, Mingyang Su, You Zhou, Naoya Iwamoto, Bo Zheng, and Michael~J Black.
\newblock Emage: Towards unified holistic co-speech gesture generation via masked audio gesture modeling.
\newblock \emph{arXiv preprint arXiv:2401.00374}, 2023.

\bibitem[Liu et~al.(2022{\natexlab{c}})Liu, Wu, Zhou, Xu, Qian, Lin, Zhou, Wu, Dai, and Zhou]{liu2022learning}
Xian Liu, Qianyi Wu, Hang Zhou, Yinghao Xu, Rui Qian, Xinyi Lin, Xiaowei Zhou, Wayne Wu, Bo Dai, and Bolei Zhou.
\newblock Learning hierarchical cross-modal association for co-speech gesture generation.
\newblock In \emph{{Computer Vision and Pattern Recognition (CVPR)}}, pages 10462--10472, 2022{\natexlab{c}}.

\bibitem[Ma et~al.(2023)Ma, Cao, Yi, Zhang, and Tao]{ma2023grammar}
Sihan Ma, Qiong Cao, Hongwei Yi, Jing Zhang, and Dacheng Tao.
\newblock Grammar: Ground-aware motion model for 3d human motion reconstruction.
\newblock In \emph{Proceedings of the 31st ACM International Conference on Multimedia}, pages 2817--2828, 2023.

\bibitem[Ma et~al.(2024)Ma, Cao, Zhang, and Tao]{ma2024richcat}
Sihan Ma, Qiong Cao, Jing Zhang, and Dacheng Tao.
\newblock Contact-aware human motion generation from textual descriptions.
\newblock \emph{arXiv preprint arXiv:2403.15709}, 2024.

\bibitem[Maas et~al.(2013)Maas, Hannun, Ng, et~al.]{maas2013rectifier}
Andrew~L Maas, Awni~Y Hannun, Andrew~Y Ng, et~al.
\newblock Rectifier nonlinearities improve neural network acoustic models.
\newblock In \emph{{International Conference on Machine Learning (ICML)}}, page~3, 2013.

\bibitem[Mao et~al.(2022)Mao, Liu, and Salzmann]{mao2022weakly}
Wei Mao, Miaomiao Liu, and Mathieu Salzmann.
\newblock Weakly-supervised action transition learning for stochastic human motion prediction.
\newblock In \emph{{Computer Vision and Pattern Recognition (CVPR)}}, pages 8151--8160, 2022.

\bibitem[Ng et~al.(2022)Ng, Joo, Hu, Li, Darrell, Kanazawa, and Ginosar]{ng2022learning}
Evonne Ng, Hanbyul Joo, Liwen Hu, Hao Li, Trevor Darrell, Angjoo Kanazawa, and Shiry Ginosar.
\newblock Learning to listen: Modeling non-deterministic dyadic facial motion.
\newblock In \emph{{Computer Vision and Pattern Recognition (CVPR)}}, pages 20395--20405, 2022.

\bibitem[Pavlakos et~al.(2019)Pavlakos, Choutas, Ghorbani, Bolkart, Osman, Tzionas, and Black]{pavlakos2019expressive}
Georgios Pavlakos, Vasileios Choutas, Nima Ghorbani, Timo Bolkart, Ahmed~AA Osman, Dimitrios Tzionas, and Michael~J Black.
\newblock Expressive body capture: {3D} hands, face, and body from a single image.
\newblock In \emph{{Computer Vision and Pattern Recognition (CVPR)}}, pages 10975--10985, 2019.

\bibitem[Peng et~al.(2021)Peng, Liu, Xu, and Li]{peng2021generating}
Jialun Peng, Dong Liu, Songcen Xu, and Houqiang Li.
\newblock Generating diverse structure for image inpainting with hierarchical vq-vae.
\newblock In \emph{{Computer Vision and Pattern Recognition (CVPR)}}, pages 10775--10784, 2021.

\bibitem[Petrovich et~al.(2021)Petrovich, Black, and Varol]{petrovich2021action}
Mathis Petrovich, Michael~J Black, and G{\"u}l Varol.
\newblock Action-conditioned 3d human motion synthesis with transformer vae.
\newblock In \emph{{International Conference on Computer Vision ({ICCV})}}, pages 10985--10995, 2021.

\bibitem[Petrovich et~al.(2022)Petrovich, Black, and Varol]{petrovich2022temos}
Mathis Petrovich, Michael~J Black, and G{\"u}l Varol.
\newblock Temos: Generating diverse human motions from textual descriptions.
\newblock In \emph{{European Conference on Computer Vision (ECCV)}}, pages 480--497. Springer, 2022.

\bibitem[Plappert et~al.(2016)Plappert, Mandery, and Asfour]{plappert2016kit}
Matthias Plappert, Christian Mandery, and Tamim Asfour.
\newblock The kit motion-language dataset.
\newblock \emph{Big data}, 4\penalty0 (4):\penalty0 236--252, 2016.

\bibitem[Poggi et~al.(2005)Poggi, Pelachaud, Rosis, Carofiglio, and Carolis]{poggi2005greta}
Isabella Poggi, Catherine Pelachaud, F~de Rosis, Valeria Carofiglio, and B~De Carolis.
\newblock Greta. {A} believable embodied conversational agent.
\newblock In \emph{Multimodal Intelligent Information Presentation}, pages 3--25. 2005.

\bibitem[Qi et~al.(2023)Qi, Liu, Li, Hou, Xin, and Yu]{qi2023emotiongesture}
Xingqun Qi, Chen Liu, Lincheng Li, Jie Hou, Haoran Xin, and Xin Yu.
\newblock Emotiongesture: Audio-driven diverse emotional co-speech 3d gesture generation.
\newblock \emph{arXiv preprint arXiv:2305.18891}, 2023.

\bibitem[Qin et~al.(2022)Qin, Zheng, and Zhou]{qin2022motion}
Jia Qin, Youyi Zheng, and Kun Zhou.
\newblock Motion in-betweening via two-stage transformers.
\newblock \emph{{Transactions on Graphics (TOG)}}, 41\penalty0 (6):\penalty0 1--16, 2022.

\bibitem[Razavi et~al.(2019)Razavi, Van~den Oord, and Vinyals]{razavi2019generating}
Ali Razavi, Aaron Van~den Oord, and Oriol Vinyals.
\newblock Generating diverse high-fidelity images with vq-vae-2.
\newblock \emph{{Conference on Neural Information Processing Systems (NeurIPS)}}, 32, 2019.

\bibitem[Siyao et~al.(2022)Siyao, Yu, Gu, Lin, Wang, Qian, Loy, and Liu]{siyao2022bailando}
Li Siyao, Weijiang Yu, Tianpei Gu, Chunze Lin, Quan Wang, Chen Qian, Chen~Change Loy, and Ziwei Liu.
\newblock Bailando: {3D} dance generation by actor-critic {GPT} with choreographic memory.
\newblock In \emph{{Computer Vision and Pattern Recognition (CVPR)}}, pages 11050--11059, 2022.

\bibitem[Van Den~Oord et~al.(2017)Van Den~Oord, Vinyals, et~al.]{van2017neural}
Aaron Van Den~Oord, Oriol Vinyals, et~al.
\newblock Neural discrete representation learning.
\newblock In \emph{{Conference on Neural Information Processing Systems (NeurIPS)}}, 2017.

\bibitem[Vaswani et~al.(2017)Vaswani, Shazeer, Parmar, Uszkoreit, Jones, Gomez, Kaiser, and Polosukhin]{vaswani2017attention}
Ashish Vaswani, Noam Shazeer, Niki Parmar, Jakob Uszkoreit, Llion Jones, Aidan~N Gomez, {\L}ukasz Kaiser, and Illia Polosukhin.
\newblock Attention is all you need.
\newblock In \emph{{Conference on Neural Information Processing Systems (NeurIPS)}}, 2017.

\bibitem[Wang et~al.(2021)Wang, Xu, Xu, Liu, and Wang]{wang2021synthesizing}
Jiashun Wang, Huazhe Xu, Jingwei Xu, Sifei Liu, and Xiaolong Wang.
\newblock Synthesizing long-term 3d human motion and interaction in 3d scenes.
\newblock In \emph{{Computer Vision and Pattern Recognition (CVPR)}}, pages 9401--9411, 2021.

\bibitem[Wu and Flierl(2019)]{wu2019learning}
Hanwei Wu and Markus Flierl.
\newblock Learning product codebooks using vector-quantized autoencoders for image retrieval.
\newblock In \emph{2019 IEEE Global Conference on Signal and Information Processing (GlobalSIP)}, pages 1--5. IEEE, 2019.

\bibitem[Xing et~al.(2023)Xing, Xia, Zhang, Cun, Wang, and Wong]{xing2023codetalker}
Jinbo Xing, Menghan Xia, Yuechen Zhang, Xiaodong Cun, Jue Wang, and Tien-Tsin Wong.
\newblock Codetalker: Speech-driven 3d facial animation with discrete motion prior.
\newblock In \emph{{Computer Vision and Pattern Recognition (CVPR)}}, pages 12780--12790, 2023.

\bibitem[Yang et~al.(2023)Yang, Wu, Li, Zhang, Hao, Bao, Cheng, and Xiao]{yang2023diffusestylegesture}
Sicheng Yang, Zhiyong Wu, Minglei Li, Zhensong Zhang, Lei Hao, Weihong Bao, Ming Cheng, and Long Xiao.
\newblock Diffusestylegesture: Stylized audio-driven co-speech gesture generation with diffusion models.
\newblock In \emph{Proceedings of the International Joint Conference on Artificial Intelligence (IJCAI-23)}, pages 5860--5868, 2023.

\bibitem[Yazdian et~al.(2022)Yazdian, Chen, and Lim]{yazdian2022gesture2vec}
Payam~Jome Yazdian, Mo Chen, and Angelica Lim.
\newblock {Gesture2Vec}: Clustering gestures using representation learning methods for co-speech gesture generation.
\newblock In \emph{{International Conference on Intelligent Robots and Systems (IROS)}}, pages 3100--3107. IEEE, 2022.

\bibitem[Yi et~al.(2023)Yi, Liang, Liu, Cao, Wen, Bolkart, Tao, and Black]{yi2023generating}
Hongwei Yi, Hualin Liang, Yifei Liu, Qiong Cao, Yandong Wen, Timo Bolkart, Dacheng Tao, and Michael~J Black.
\newblock Generating holistic 3d human motion from speech.
\newblock In \emph{Proceedings of the IEEE/CVF Conference on Computer Vision and Pattern Recognition}, pages 469--480, 2023.

\bibitem[Yin et~al.(2023)Yin, Wang, He, Liu, Zhao, Li, Jin, and Lin]{yin2023emog}
Lianying Yin, Yijun Wang, Tianyu He, Jinming Liu, Wei Zhao, Bohan Li, Xin Jin, and Jianxin Lin.
\newblock Emog: Synthesizing emotive co-speech 3d gesture with diffusion model.
\newblock \emph{arXiv preprint arXiv:2306.11496}, 2023.

\bibitem[Yoon et~al.(2020)Yoon, Cha, Lee, Jang, Lee, Kim, and Lee]{yoon2020speech}
Youngwoo Yoon, Bok Cha, Joo-Haeng Lee, Minsu Jang, Jaeyeon Lee, Jaehong Kim, and Geehyuk Lee.
\newblock Speech gesture generation from the trimodal context of text, audio, and speaker identity.
\newblock \emph{{Transactions on Graphics (TOG)}}, 39\penalty0 (6):\penalty0 1--16, 2020.

\bibitem[Zhang et~al.(2023)Zhang, Zhang, Cun, Huang, Zhang, Zhao, Lu, and Shen]{zhang2023t2m}
Jianrong Zhang, Yangsong Zhang, Xiaodong Cun, Shaoli Huang, Yong Zhang, Hongwei Zhao, Hongtao Lu, and Xi Shen.
\newblock T2m-gpt: Generating human motion from textual descriptions with discrete representations.
\newblock In \emph{{Computer Vision and Pattern Recognition (CVPR)}}, 2023.

\bibitem[Zhang et~al.(2024)Zhang, Cai, Pan, Hong, Guo, Yang, and Liu]{zhang2022motiondiffuse}
Mingyuan Zhang, Zhongang Cai, Liang Pan, Fangzhou Hong, Xinying Guo, Lei Yang, and Ziwei Liu.
\newblock Motiondiffuse: Text-driven human motion generation with diffusion model.
\newblock \emph{{Transactions on Pattern Analysis and Machine Intelligence (TPAMI)}}, 2024.

\bibitem[Zhou et~al.(2019)Zhou, Barnes, Lu, Yang, and Li]{zhou2019continuity}
Yi Zhou, Connelly Barnes, Jingwan Lu, Jimei Yang, and Hao Li.
\newblock On the continuity of rotation representations in neural networks.
\newblock In \emph{{Computer Vision and Pattern Recognition (CVPR)}}, pages 5745--5753, 2019.

\bibitem[Zhu et~al.(2023{\natexlab{a}})Zhu, Liu, Liu, Qian, Liu, and Yu]{zhu2023taming}
Lingting Zhu, Xian Liu, Xuanyu Liu, Rui Qian, Ziwei Liu, and Lequan Yu.
\newblock Taming diffusion models for audio-driven co-speech gesture generation.
\newblock In \emph{{Computer Vision and Pattern Recognition (CVPR)}}, pages 10544--10553, 2023{\natexlab{a}}.

\bibitem[Zhu et~al.(2023{\natexlab{b}})Zhu, Ma, Ro, Ci, Zhang, Shi, Gao, Tian, and Wang]{zhu2023human}
Wentao Zhu, Xiaoxuan Ma, Dongwoo Ro, Hai Ci, Jinlu Zhang, Jiaxin Shi, Feng Gao, Qi Tian, and Yizhou Wang.
\newblock Human motion generation: A survey.
\newblock \emph{{Transactions on Pattern Analysis and Machine Intelligence (TPAMI)}}, 2023{\natexlab{b}}.

\bibitem[Zhuang et~al.(2022)Zhuang, Wang, Chai, Wang, Shao, and Xia]{zhuang2022music2dance}
Wenlin Zhuang, Congyi Wang, Jinxiang Chai, Yangang Wang, Ming Shao, and Siyu Xia.
\newblock Music2dance: Dancenet for music-driven dance generation.
\newblock \emph{ACM Transactions on Multimedia Computing, Communications, and Applications (TOMM)}, 18\penalty0 (2):\penalty0 1--21, 2022.

\end{thebibliography}
\bibliographystyle{plainnat}

\clearpage
\newpage
\begin{appendices} 
\label{appendices}
\section{Method Details}

\subsection{PQ-VAE}
The PQ-VAE processes motion sequences, denoted as $M_{1:N}$, utilizing an encoder-decoder architecture tailored for temporal data \cite{yi2023generating}. The encoder comprises four residual blocks, each featuring a series of three temporal convolution layers. These layers are parameterized with kernel size, stride, and padding set to 3, 1, and 1, respectively, and are followed by batch normalization \cite{ioffe2015batch} and Leaky ReLU activation \cite{maas2013rectifier} for non-linear transformations. Additionally, temporal convolutions (kernel size 4, stride 2, padding 1) are placed between residual blocks to adjust the temporal resolution, setting the temporal window $w$ to 8. A fully-connected layer precedes the quantization step, serving dimensionality reduction. The decoder mirrors the encoder's architecture, ensuring symmetry in information reconstruction.

\subsection{Predictor}
The Predictor leverages dual condition encoders and a transformer-based decoder. The audio encoder employs 3 temporal convolution layers (kernel size 4, stride 2, padding 1), followed by batch normalization \cite{ioffe2015batch} and Leaky ReLU activation \cite{maas2013rectifier}, focusing on audio feature extraction. In contrast, the motion context encoder utilizes 10 gated convolution layers, catering to motion context comprehension \cite{peng2021generating, ding2018trunk}. The transformer-based decoder consists of an embedding layer and six decoder blocks, each integrating a self-attention layer, a cross-attention layer, and a linear layer, followed by an AdaIN layer for style normalization. A fully-connected layer finalizes the structure, adjusting output dimensions to match the target motion specifications.

\subsection{Refiner}
Employing a transformer-based architecture akin to the Predictor's decoder (excluding the embedding layer), the Refiner fine-tunes motion predictions. During training phase, input to the Refiner combines ground truth (GT) motion $M^{gt}_{1:N}$ and PQ-reconstructed motion $M^{pq}_{1:N}$ via:
\begin{equation}
\overline{M}_{1:N} = I \odot M^{gt}_{1:N} + (1 - I) \odot M^{pq}_{1:N},
\end{equation}
where $\odot$ denotes element-wise multiplication. The Refiner outputs refined motion $M^{r}_{1:N}$, guided by input audio $A_{1:N}$, mask $I$ and speaker identity $D_{1:N}$:
\begin{equation}
{M}^{r}_{1:N} = \text{Refiner}(\overline{M}_{1:N}; A_{1:N}, I, D_{1:N}).
\end{equation}
The Refiner is optimized using the loss function $\mathcal{L}_{refine}$, which is formulated as follow:
\begin{equation}
\small
\mathcal{L}_{refine} = \mathcal{L}_{1}(I \odot M^{gt}_{1:N}, I \odot M^{r}_{1:N}) + \mathcal{L}_{1}(V^{gt}_{1:N-1}, V^{r}_{1:N-1}).
\end{equation}
Here, $\mathcal{L}_{1}$ is L1 reconstruction loss, while $V^{gt}_{1:N-1}$ and $V^{r}_{1:N-1}$ refer to the velocity of GT and generated holistic body motion, respectively.

\section{More Comparison}

\begin{table}
\centering
\resizebox{0.9\linewidth}{!}{
\begin{tabular}{llll}
\toprule
Method              & FGD $\downarrow$ & MAE $\downarrow$       & BC (GT=0.847) \\ \hline
Habibie et al.      & 44.60 & 8.59      & 0.964 (GT+0.117) \\
TalkSHOW            & 6.60 & 9.39      & 0.885 (GT+0.038) \\
ProbTalk (Ours)     & 3.98 & 7.79      & 0.818 (GT-0.029) \\ \hline
\end{tabular}
}
\caption{More metrics.}
\label{tab:metrics_cr}
\end{table}

\paragraph{More Metrics.}
We add the \textbf{Mean Absolute Error} (MAE) to quantitatively assess the difference between the ground truth and the motion generated by our model. Additionally, we introduced the \textbf{Beat Consistency} (BC) metric to evaluate the synchronization between the generated motion and the corresponding audio. The outcomes of these evaluations are presented in Tab. \ref{tab:metrics_cr}. The superior performance in both the MAE and BC metrics demonstrates that our model's generated outputs exhibit the highest degree of fidelity to the ground truth.

\begin{figure}
    \centering
    \includegraphics[width=0.95\columnwidth]{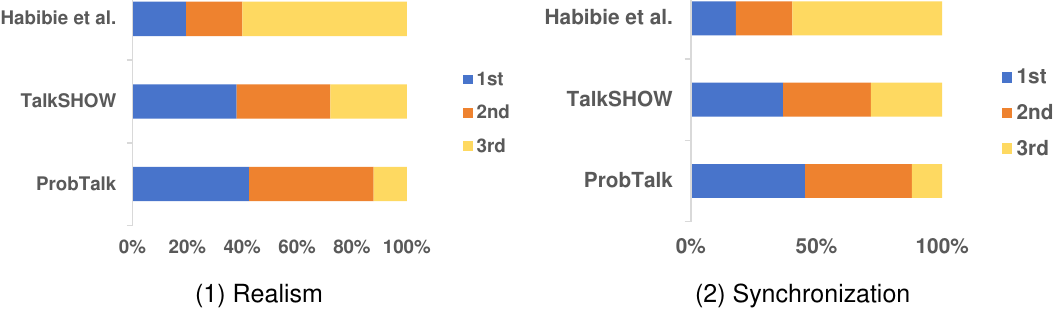}
    \caption{User study.}
    \label{fig:user_study}
\end{figure}

\paragraph{User Study.}
We conduct a user study to compare the realism and synchronization of our method with existing works. We randomly sample 20 audios. 12 participants were asked to rank videos generated by three methods based on their realism and synchronization. Results in Fig. \ref{fig:user_study} show that our method outperforms the others in both realism and synchronization.

\begin{table}[]
    \centering
    \resizebox{\columnwidth}{!}{
    \begin{tabular}{l|ccccc}
    \hline
        ~ & FGD↓ & BC & Diversity↑ & MSE↓ & LVD↓ \\ \hline
        Ground Truth & 0 & 6.856 & 13.05 & 0 & 0 \\
        FaceFormer \cite{fan2022faceformer} & - & - & - & 7.787 & 7.593 \\
        CodeTalker \cite{xing2023codetalker} & - & - & - & 8.026 & 7.766 \\ 
        S2G \cite{ginosar2019gestures} & 28.15 & 4.683 & 5.971 & - & - \\ 
        Trimodal \cite{yoon2020speech} & 12.41 & 5.933 & 7.724 & - & - \\ 
        HA2G \cite{liu2022learning} & 12.32 & 6.779 & 8.626 & - & - \\ 
        DisCo \cite{liu2022disco} & 9.417 & 6.439 & 9.912 & - & - \\ 
        CaMN \cite{liu2022beat} & 6.644 & 6.769 & 10.86 & - & - \\ 
        DiffStyleGesture \cite{yang2023diffusestylegesture} & 8.811 & 7.241 & 11.49 & - & - \\ 
        Habibie et al. \cite{habibie2021learning} & 9.040 & 7.716 & 8.213 & 8.614 & 8.043 \\ 
        TalkShow \cite{yi2023generating} & 6.209 & 6.947 & 13.47 & 7.791 & 7.771 \\ 
        EMAGE \cite{liu2023emage}    & 5.512 & 7.724 & 13.06 & 7.680 & 7.556 \\ 
        ProbTalk (Ours) & 6.170 & 8.099 & 10.43 & 8.990 & 8.385 \\ \hline
    \end{tabular}
    }
    \caption{Quantitative evaluation on BEAT-X.}
    \label{tab:beatv2}
\end{table}
\begin{figure*}[ht]
    \centering
    \includegraphics[width=\linewidth]{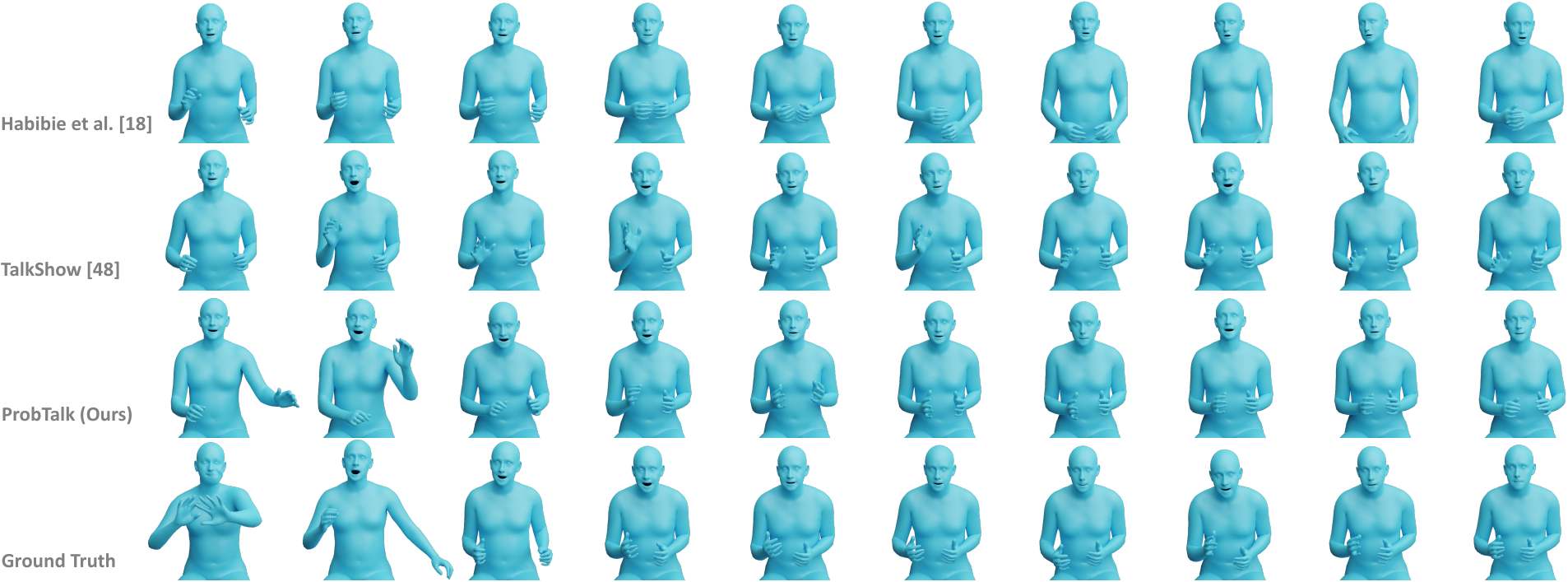}
    \caption{Motion sequences selected based on their minimal Mean Squared Error (MSE), incorporating symmetry by treating movements of corresponding body parts (e.g., left and right hands) as equivalent in the evaluation.}
    \label{fig:acc}
\end{figure*}

\paragraph{Experiments on Beat-X Dataset.}
Our experimental evaluation conducted on the Beat-X dataset \cite{liu2023emage} is shown in Tab. \ref{tab:beatv2}. We follow the experimental configuration from \cite{liu2023emage}, employing data from \textbf{Speaker 2} for model training and validation. This ensures a direct comparison with benchmarks and aligns with prior research. However, it is important to acknowledge the potential for \textbf{overfitting} due to the limited data. As such, the outcomes of these experiments should be interpreted as indicative rather than definitive. 

\paragraph{More Qualitative Comparison.} 
To further validate the accuracy of the generated motions, we generate 32 samples using each method. Subsequently, we select the samples exhibiting the minimal Mean Squared Error (MSE) for each method. The corresponding results are presented in Fig. \ref{fig:acc}. The results demonstrate that our method generates a closer approximation to the true motion sequences, thus highlighting the enhanced accuracy of our generated samples.

\end{appendices}


\end{document}